\documentclass[12pt,letterpaper]{article}

\def\In{{\mbox {\rm in}}}

\newtheorem{proposition}{Proposition}[section]
\newtheorem{definition}{Definition}[section]

\usepackage{times}
\usepackage{epsfig}
\usepackage{graphicx}
\usepackage{amsmath}
\usepackage{amssymb}
\usepackage{subfigure}
\usepackage{multirow}
\usepackage{slashbox}
\usepackage{array}

\begin{document}

\title{A New Solution to the Relative Orientation Problem using only 3 Points and the Vertical Direction}

\author{Mahzad Kalantari\\
ENSG, Institut G\'eographique National-France\\
IVC Lab, Institut Recherche Communications\\
 Cybern\'etique de Nantes (IRCCyN) UMR CNRS 6597\\
Institution1 address\\
{\tt\small mahzad.kalantari@ensg.eu}
\and
Amir Hashemi\\
Department of Mathematical Sciences,\\
Isfahan University of Technology Isfahan-IRAN\\
{\tt\small amir.hashemi@cc.iut.ac.ir}
\and
Franck Jung\\
DDE - Seine Maritime, France\\
{\tt\small Franck.Jung@equipement.gouv.fr}
\and
JeanPierre Guedon\\
IVC Lab, Institut Recherche Communications\\
Cybern\'etique de Nantes (IRCCyN) UMR CNRS 6597\\
{\tt\small jean-pierre.guedon@polytech.univ-nantes.fr}
}

\maketitle

\begin{abstract}
This paper presents a new method to recover the relative pose between two images, using three points and the vertical direction information. 
The vertical direction can be determined in two ways: 1- using direct physical measurement like IMU (inertial measurement unit),
2- using vertical vanishing point.
This knowledge of the vertical direction solves 2 unknowns among the 3 parameters of the relative rotation, so that only 3 homologous points are requested to position a couple of images. Rewriting the coplanarity equations leads to a simpler solution.
The remaining unknowns resolution is performed by an algebraic method using  Gr\"obner bases. The elements necessary to build a specific algebraic solver are given in this paper, allowing for a real-time implementation. The results on real and synthetic data show the efficiency of this  method.
\end{abstract}


\section{Introduction}
This paper presents an efficient solution to the relative orientation problem in calibration setting.
In such a situation, the intrinsic parameters of the camera, e.g. the focal length, the camera distortion are assumed to be  a priori known. In this case the relative orientation linking two views is modeled by 5 unknowns: the rotation matrix (3 unknowns) and the translation (2 unknowns up to a scale). Its resolution using only five points, in a direct and fast way, has been considered as a major research subject since the eighties \cite{Philip5points} up to now \cite{Triggs5points}, \cite{Nister04}, \cite{stewenius-engels-etal-ijprs-06}, \cite{LiHartleyfiveEasy}, \cite{BatraFiveAlernative}, \cite{Kukelovap5td}.
In this paper we use the knowledge of the vertical direction to solve the relative orientation problem for two reasons: \\
1- the increased use of MEMS-IMU (inertial measurement unit) in electronic personal devices such as smart phones, digital cameras and the low price IMU.  The sensors fusion (camera-IMU) is not the goal of this paper, as many authors have shown the advantage of coupling them \cite{SensorIMU}. In MEMS-IMU the accuracy of heading (rotation around the vertical axis Z) is worse than for pitch (rotation around X axis) and roll (rotation around Y axis), due to the strength of the gravity field, which has no effect on a rotation around the vertical axis. Thus the new method presented in this paper takes a considerable benefit from a combination of data from MEMS-IMU and from use of 3 homologous points, that strengthen the very weakness of IMU data.\\
2- today very performant algorithms based on image analysis are available, that allow to calculate the vertical direction with high accuracy.  If we have only a set of calibrated images we can also determine the vertical direction using vanishing points extraction. A lot of algorithms \cite{Barnard}, \cite{Lutton}, \cite{Shufelt}, on such topics exist in the literature. These algorithms are very useful in urban and man-made environments  \cite{Heuvel}, \cite{Teller}, \cite{PlanarGrouping}, \cite{Kalantari}.\\
The use of the vertical direction so as to reduce the disparity between two frames, to simplify 3D vision, has already been considered by \cite{Vieville}. But most papers use a fixed stereoscopic baseline, and here we consider that we have no knowledge about it. Furthermore, most paper \cite{Vieville} try to solve the problem using iterative methods or non minimal settings (e.g. more than three points).\\
\section{Our contribution to the relative orientation problem}
The main contribution of this paper is to provide an efficient algorithm to estimate the relative orientation using the vertical direction as an external information in the minimal case, using 3 points.
Once the vertical direction is defined, we inject this information in relative orientation, based on coplanarity equation.  
The knowledge of the vertical direction removes 2 degrees of freedom to the problem of the relative orientation. Therefore it will be enough to have only 3 homologous couples of points to solve for the 3 other unknowns: two parameters of the baseline because it is up to a scale and the angle of rotation around the vertical axis. These coplanarity constaints can be written as a system of polynomial equations.  Hence, we solve these equations using the Gr\"obner bases in a direct way. The possibility to build a solution with only 3 points is an obvious advantage in terms of computation time, in particular when sorting the undesirable solutions by classic robust estimators such as Ransac (RANdom SAmple Consensus)\cite{RanSac}. In the Section~\ref{Experiments} we show that the new 3-point method provides better accuracy and robustness to noise on relative orientation estimation.\\
The paper is organized as follows. In the section \ref{geometryUse} we present the geometric framework of our system. Section \ref{geometryEquaVP} rewrites the coplanarity constraint using the vertical direction knowledge. The resolution of polynomial system with the help of  Gr\"obner bases is described in Section \ref{Groebner}.
The assessment  of the algorithm in noisy conditions is studied in Section \ref{sec:noise}, where the 3-point algorithm is compared to the well known 5-point algorithm. In Section \ref{sec:realperf} a comparaison with real image database is performed.

\section{Coordinate systems and geometry framework}\label{geometryUse}
\begin{figure}
\begin{center}
\includegraphics[width=0.5\linewidth]{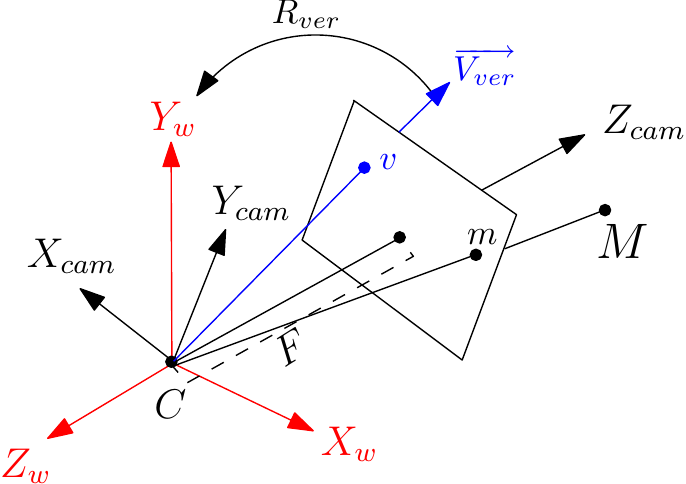}
\end{center}
   \caption{Coordinate systems and geometry overview. The vector  ${V_{ver}}$ is the vector of vertical vanishing point and pierces the image plane in $v$. $R_{ver}$ is define in Section \ref{absoluOrientation}}
\label{figsystemeOfCoordinate}
\end{figure}

The classical coordinate system of camera (cf. figure~\ref{figsystemeOfCoordinate}) used in computer vision has been chosen \cite{HartleyMultipleView}. In this camera system $(X_{cam},\ Y_{cam},\ Z_{cam}) $, the focal plane is $Z_{cam} = F$ , F being the focal length.
Given the calibration matrix $K$ (a 3x3 matrix that includes the information of focal length, skew of the camera, etc.), the view is normalized by transforming all points by the inverse of $K$, $ \hat{m} = K^{-1}m$, in which $m$ is a 2-coordinates point in the image. Thus the new calibration matrix of the view becomes the identity matrix. $M$ is the object point. In the rest of the paper we suppose that all image 2D-coordinates of the point are normalized.
For a stereo system in relative orientation, the center of the world space coordinate system is the optical center $C$ of the left image, with the same directions of axes.
The world coordinate system is denoted by $(X_{w},\ Y_{w},\ Z_{w})$. In this system the $Y_w$ axis is along the physical vertical of the world space.

\section{Using the vertical direction knowledge for relative orientation}\label{geometryEquaVP}
\subsection{Use the IMU information}
If we have of an IMU coupled with the camera, we need only to know the rotation angle ($\alpha$) around X axis and Z axis ($\gamma$) based on our coordinates system. So the rotation matrix equals:
\begin{equation}\label{rotaIMU}
R_{ver}= \begin{bmatrix} \cos \gamma & -\sin \gamma & 0 \\ \sin \gamma & \cos \gamma & 0 \\ 0 & 0 & 1 \end{bmatrix} \begin{bmatrix} 1 & 0 & 0 \\ 0 & \cos \alpha  & -\sin \alpha  \\ 0 & \sin \alpha  & \cos \alpha  \end{bmatrix} 
\end{equation}
\subsection{Use the information given by vertical vanishing point}\label{absoluOrientation}
If we only have a set of calibrated images of a man-made environment we can extract the vertical direction using vertical vanishing point.
Let us suppose that $\overrightarrow{V_{ver}}$ be the vector joining $C$ to the vanishing point in the image plane expressed in the camera system, and $\overrightarrow{Y_w} (0\ , 1\ , 0)$ be the $Y$ axis of the world system ((see figure \ref{figsystemeOfCoordinate}).
We perform the rotation that transforms $\overrightarrow{V_{ver}}$ into $\overrightarrow{Y_w}$. Thus, we determine the rotation  axis $\overrightarrow{\mathbf{\omega}}$ and the rotation angle $\theta$ in the following way: $\overrightarrow{\mathbf{\omega}} = \overrightarrow{V_{ver}} \otimes \overrightarrow{Y_w}$, after simplification and normalisation $\overrightarrow{\mathbf{\omega}} = [\frac{V_z}{d}\ , 0\ ,\frac{-V_x}{d}]$, where $d = \sqrt{V_z^2 + V_x^2}$ ,
$\theta = \arccos{(\overrightarrow{V_{ver}} \cdot \overrightarrow{Y_w})}$, so after simplification, $\theta = \arccos{(V_y)}$.
Using Olinde-Rodrigues formula we get the following rotation matrix :
\begin{equation}\label{olinde}
R_{ver} = I \cos\theta + \sin\theta \left[ \mathbf{\omega} \right]_\times + (1 - \cos\theta) \mathbf{\omega}\ ^t\mathbf{\omega} .
\end{equation}
The rotation $(R_{ver}$) given by equation \ref{rotaIMU} or \ref{olinde} is then applied to all 2D points obtained in each image, $\hat{m}$ is replaced by $R_{ver}\hat{m}$.
\subsection{Rewriting the coplanarity constraint}\label{sysPoly}
First, we recall that for a pair of homologous points $\hat{m^1}$ and $\hat{m^2}$ of a pinhole camera, the constraint on these 2 points is expressed by the equation of coplanarity:
\begin{equation}\label{essentialMatrix}
\begin{bmatrix}\hat{m_{x}^{2}}&\hat{m_{y}^{2}}&1\end{bmatrix}E\begin{bmatrix}\hat{m_{x}^{1}}\\\hat{m_{y}^{2}}\\1\end{bmatrix}=0.
\end{equation}
where $E$ is a 3x3 rank-2 essential matrix \cite{HartleyMultipleView}. We can also express this constraint by the equation \ref{algebricCoplanarity}.
\begin{equation}\label{algebricCoplanarity}
\begin{bmatrix}\hat{m_{x}^{2}}&\hat{m_{y}^{2}}&1\end{bmatrix}\begin{bmatrix}\,0&\!T_z&\,\,\,-T_y\\\,\,\,-T_z&\,0&\!T_x\\T_y&\,\,-T_x&\,0\end{bmatrix}R\begin{bmatrix}\hat{m_{x}^{1}}\\\hat{m_{y}^{2}}\\1\end{bmatrix}=0.
\end{equation}
However, if we apply the rotation $(R_{ver})$ obtained in equation \ref{olinde} to all homologous points, before we take in account this constraint (equation \ref{algebricCoplanarity}), the rotation R is expressed in a simpler way, as it remains only one parameter of rotation to estimate, the angle $\phi$ around the $Y$ axis (vertical axis). Thus: 
\begin{equation}\label{RotaY}
R_{\phi}=\begin{bmatrix}
         \cos{\phi} &0& -\sin{\phi} \\
           0 &1& 0\\
           \sin{\phi} &0& \cos{\phi} \\
\end{bmatrix}
\end{equation}
Using $t = \tan{\frac{\phi}{2}}$, we replace $\cos{\phi}$ by ${(1 - t^{2})}/{(1 + t^2)}$ and $\sin{\phi}$ by ${2t}/{(1 + t^2)}$.
The new coplanarity equation is rewritten as:
\begin{equation}\label{NewCopla}
\begin{split}
(-2&\hat{m_{x}^{2}}T_yt+\hat{m_{y}^{2}}(T_z(1-t^2)+2T_xt)-\\
&\hat{m_{z}^{1}}T_y(1-t^2))\hat{m_{x}^{1}}+(\hat{m_{x}^{2}}(1+t^2)T_z+\\
&\hat{m_{z}^{2}}(1+t^2)T_x)\hat{m_{y}^{1}}+(\hat{m_{x}^{2}}T_y(1t^2)+\\
&\hat{m_{y}^{2}}(2T_zt-T_x(1-t^2))-2\hat{m_{z}^{2}}T_yt)\hat{m_{z}^{1}} =0.
\end{split}
\end{equation}

3 pairs of homologous points allows for instancing equation 6 as {$\{f_2,f_3,f_4\}$ with remaining unknowns $T_x, T_y, T_z$ and $t$. The corresponding base is only composed from two degree of freedom since no scale modeling has been yet performed.
Therefore it is necessary either to fix a component of the base to 1, either to add the constraint of normality. We choose this last one: $f_1 \equiv T_x^2 + T_y^2 + T_z^2 - 1 = 0$. The advantage is that it allows to get a more general modeling.
We have therefore a system of 4 polynomial equations of degree 3 $\{f_1,f_2,f_3,f_4\}$. Now we describe the direct resolution of this polynomial system using the Gr\"{o}bner bases.

\section{Resolution of the relative orientation equation using Gr\"obner bases} \label{Groebner}
We recall first the basic definitions of Gr\"obner bases, and also the link between Gr\"obner bases and linear algebra. Then, we use these concepts to derive a specific algorithm to compute the Gr\"obner basis of the system of polynomials defined in Section \ref{sysPoly}.
\subsection{Properties of Gr\"obner basis}
The notion of Gr\"obner basis was introduced by B. Buchberger, who gave the first algorithm to compute it (see \cite{Bruno2}). This algorithm is implemented in most general computer algebra systems like {\sc Maple}, {\sc Mathematica}, {\sc Singular} \cite{Singular}, {\sc Macaulay2} \cite{Macaulay}, {\sc Cocoa} \cite{Cocoa} and {\sc Salsa} software \cite{Salsa}.
Let $R=K[x_1,\ldots ,x_{n}]$ be a polynomial ring where $K$ is an
arbitrary field. Let $f_1,\ldots ,f_k\in R$ be a sequence of $k$
polynomials and let $I=\langle f_1,\ldots ,f_k \rangle$ be an
ideal of $R$ generated by the $f_i$'s. We need also a monomial
ordering on $R$. We recall here the definition of the {\it degree
reverse lexicographic ordering} (DRL), denoted by $\prec$, which is an especial monomial ordering having some interesting computational properties. For this we denote respectively by $\deg(m)$ (resp. $\deg_i(m)$) the total
degree (resp. the degree in $x_i$) of a monomial $m$. If $m$ and $m'$
are monomials, then $m\prec m'$ if and only if the last non zero
entry in the sequence
$(\deg_1(m')-\deg_1(m),\ldots,\deg_n(m')-\deg_n(m) ,
\deg(m)-\deg(m') )$ is negative (see \cite{little}).

Let $\In(f)\in R$ be the initial (greatest) monomial of a polynomial $f\in R$ with respect to $\prec$ and
 $\In(I)=\langle \In(f) \ | \ f\in I \rangle$ be the initial ideal of $I$.
\begin{definition}[Gr\"obner basis]
 A finite subset $G\subset I$ is a Gr\"obner basis of $I$ w.r.t. $\prec$ if $\langle \In(G) \rangle=\In(I)$.
\end{definition}
\begin{definition}[Reduced Gr\"obner basis]
 A  Gr\"obner basis $G$ of $I$ is called reduced if for all $g\in G$, $g$ is monic and no monomial of $g$ lies in $\langle \In(G\setminus \{g \}) \rangle $.
\end{definition}
\begin{proposition}[\cite{little}, Proposition $6$, page $92$]
 Every ideal has a unique reduced  Gr\"obner basis.
\end{proposition}
\subsection{Macaulay matrix}
We recall now the definition of a Macaulay matrix and we explain who we could use it to compute the Gr\"obner basis of an ideal. With the notations of above subsection, we consider the ideal $I$ generated by the $f_i$'s and $\prec$ be DRL monomial ordering. We suppose that we know the maximum degree $d$ of monomials which appear in the representation of the elements of the Gr\"obner basis of $I$ in terms of the $f_i$'s (in Subsection \ref{subMac}, we show how to compute such a degree for the ideal generated by polynomials defined in Subsection \ref{sysPoly}). Note that this degree is the maximum degree of monomials which appear in the computation of the Gr\"obner basis of $I$.

We can build the {\em Macaulay} matrix $M_d(f_1,\ldots ,f_k)$
(for short we denote it by $M_d$) as follows: Write down
horizontally all the monomials of degree at most $d$, ordered
following $\prec$ (the first one being the largest one). Hence,
each column of the matrix is indexed by a monomial of degree at
most $d$. Multiply each $f_i$ from $1$ to $k$ by any monomial $m$
of degree at most $d-\deg(f_i)$, and write the coefficients of
$mf_i$ under their corresponding monomials, thus giving a row of
the matrix. The rows are ordered: row $mf_i$ is before $uf_j$ if
either $i<j$ or $i=j$ and $m\prec u$.

$$
M_d= \ \bordermatrix{
 & & \text{monomials} & \text{of} & \text{degree \ at \ most}  & d &  \cr
\vdots &  &  &   &   &  &    \cr
mf_i &  &  &   &   &  &  \cr
\vdots &  &  &   &   &  &    \cr
}
$$

For any row in the matrix, consider the monomial indexing the
first non-zero column of this row. It is called the {\em leading
monomial} of the row, and is the leading monomial of the
corresponding polynomial.

{\em Gaussian elimination} applied on this matrix leads to a
Gr\"obner basis of $I$ (see \cite{Daniel83}). Indeed, call
$\tilde{M}_d$ the Gaussian elimination form of $M_d$, such that
the only elementary operation allowed for one row is the addition of  a linear
combination of the previous rows. Now, consider all the
polynomials corresponding to a row whose leading term is not the
same   in $M_d$ and $\tilde{M}_d$, then the set of these
polynomials is a Gr\"obner basis of $I$.

\subsection{Constructing the specific Macaulay matrix}
\label{subMac}
In this subsection we describe a general algorithm to compute the
Gr\"obner basis of  the system  of polynomials defined in Subsection \ref{sysPoly}. It is worth noting that when the coordinates of the input
points change, only the coefficients of polynomials change. Thus,
using Lazard's approach (see the above subsection), we build a
Macaulay matrix (and we may compute it directly when the
coordinates of the input points change), and a Gaussian
elimination on this matrix gives the Gr\"obner basis of the ideal.

Let $f_1,\ldots,f_4\in \mathbb{C}[T_x,T_y,T_z,t]$ be the system of
polynomials as defined in Subsection \ref{sysPoly}. Let $I=\langle f_1,\ldots
,f_4 \rangle$. Our first challenge is to choose a {\em good}
monomial ordering. From a good monomial ordering, we mean an
ordering for which the maximum reached degree in Gr\"obner basis
computation is minimum. Or in terms of complexity, we look for an
ordering for which the computation has the optimal complexity.
We choose DRL ordering because it typically provides for the fastest Gr\"obner basis computations.
Let us consider
DRL$(T_x,T_y,T_z,t)$. We compute first the maximum degree of monomials
which appear in the computation of the Gr\"obner basis of $I$
w.r.t. this ordering. We use this degree to study the complexity of computing Gr\"obner basis and also to construct the Macaulay matrix of $I$ to compute its Gr\"obner basis. For this, we homogenize the $f_i$'s w.r.t.
an auxiliary variable $h$ and we compute the Gr\"obner basis of
the homogenized system for DRL$(T_x,T_y,T_z,t,h)$. The maximum degree
of the elements of this basis is $6$ and therefore the maximum
degree of monomials which appear in the computation of the
Gr\"obner basis of $I$ will be $6$ (see \cite{Daniel83} for more details). 
We have tested some other monomial orderings, and it seems that this ordering is the best one. 

Our second challenge is to build $M_6(f_1,\ldots,f_4)$, say
$M$. To compute such a matrix, we have to find the  products $mf_i$,
such that a Gaussian elimination on the matrix representation of
these products leads us to the Gr\"obner basis of $I$. For this, we use the maximum reached degree in Gr\"obner basis computation which is $6$. We consider all products $mf_i$ where $m$ is a monomial of degree at most
$6-\deg(f_i)$. This gives $175$ polynomials. Among them, there
are some products which are useful to build $M$. Using the
following programme in {\sc Maple}, we could choose the useful
ones:
\begin{verbatim}
L:=NULL:
AA:=A:
for i from 1 to nops(A) do
        unassign('p'):
        X:=AA:
        member(A[i], AA, 'p'):
        AA:=subsop(p=NULL,AA):
        if IsGrobner(Macaulay(AA)) then
            L:=L,i:
        else
            AA:=X:
        fi:
od:
\end{verbatim}
 where {\tt IsGrobner} is a programme to test whether a set of
polynomials is a Gr\"obner basis for $I$ or not, and {\tt Macaulay}
is a programme which performs a Gaussian elimination on the matrix
representation of a set of polynomials. This gives $65$
polynomials of degree at most $6$. In this case, $M$
has a size $65\times 77$. Here is the list of $65$ polynomials which were
found by this way.
$$f_4, tf_4, T_zf_4, T_yf_4, T_xf_4, tT_zf_4, tT_yf_4, tT_xf_4,$$
$$T_zT_yf_4, T_zT_xf_4, T_y^2f_4, T_yT_xf_4, T_x^2f_4, tT_zT_yf_4,$$
$$tT_zT_xf_4, tT_y^2f_4, tT_yT_xf_4, tT_x^2f_4, f_3, tf_3, T_zf_3,$$
$$T_yf_3, T_xf_3, tT_zf_3, tT_yf_3, tT_xf_3, T_zT_yf_3, T_zT_xf_3,$$
$$T_y^2f_3, T_yT_xf_3, T_x^2f_3, tT_zT_xf_3, tT_y^2f_3, tT_yT_xf_3,$$
$$tT_x^2f_3, f_2, tf_2, T_zf_2, T_yf_2, T_xf_2, tT_zf_2, tT_yf_2,$$
$$tT_xf_2, T_zT_yf_2, T_zT_xf_2, T_y^2f_2, T_yT_xf_2, T_x^2f_2,$$
$$tT_zT_xf_2, tT_y^2f_2, tT_yT_xf_2, tT_x^2f_2, f_1, tf_1, T_zf_1,$$
$$T_yf_1, T_xf_1, t^2f_1, tT_yf_1, tT_xf_1, t^3f_1, t^2T_yf_1,$$
$$t^2T_xf_1, t^3T_yf_1, t^3T_xf_1$$

Remark that {\tt IsGrobner} and {\tt Macaulay} were written in {\sc Maple} and the former does {\em not} use Buchberger's criterion to test whether or not a set of polynomials is a Gr\"obner basis or not, because using this criterion is very time-consuming. In fact, we have used the properties that we can compute $\In(I)$ and a set of polynomials $G\subset I$ is a Gr\"obner basis for $I$ if $\In(G)=\In(I)$. This makes {\tt IsGrobner} very fast and efficient, and allows to do the above choice in real time.

\subsection{Constructing the specific algebraic solver}
\label{alsolver}
In this subsection,, we recall briefly an algebraic solver which uses a Gr\"obner basis to find the solutions of the system defined in Subsection \ref{sysPoly}.

Thanks to the property that the division by the ideal $I$ is well
defined when we do it w.r.t a Gr\"obner basis of $I$, we can
consider the space of all remainders on division by $I$ (see
\cite{little}). This space is called the {\em quotient ring} of
$I$, and we denote it by $A=\mathbb{C}[Tx,Ty,Tz,t]/I$. It is
well-known that if $I$ is radical then the system
$f_1=\cdots=f_4=0$ has a finite number of solutions $N$ if the
dimension of $A$ as an $\mathbb{C}$-vector space is $N$ (see
\cite{little}, Proposition $8$ page $235$). We can easily check by
the function {\tt IsRadical} of {\sc Maple} that $I$ is radical. A
basis for $A$ as a vector space is obtained from $\In(I)$ by
(\cite{little}, Theorem $6$, page $234$)
$$B=\{m \ | \ m \ \text{is a monomial and} \ m\notin \In(I)\}$$
From computing a Gr\"obner basis of $I$, we could compute
$\In(I)$, which is equal to $\In(I)=\langle  T_x, T_y,
T_z^2, t^6\rangle$ and thus the set
$$B=\{1, t, t^2, t^3, t^4, t^5, T_z, T_zt, T_zt^2, T_zt^3, T_zt^4, T_zt^5\}$$
is a basis for $A$ as an $\mathbb{C}$-vector space. Therefore, we
can conclude that the system $f_1=\cdots=f_4=0$ has $12$ solutions.
Note that we have obtained these results for an especial 
coordinates of input points. We can discuss mathematically the correctness of
these results for any set of points. But, that is out of the subject of this paper and the scope of this conference. 
We recall here briefly the eigenvalue method that we have used to solve the system $f_1=\cdots =f_4=0$, see \cite{Cox}, page $56$ for more details. For any  $f\in \mathbb{C}[T_x,T_y,Tz,t]$ let us denote by  $[f]$ the coset of $f$ in $A$. We define $m_f : A \longrightarrow A$ by the following rule:
$$m_f([g])=[f].[g]=[fg]\in A$$
Since, the ideal generated by the $f_i$'s is zero-dimensional, then $A$ is a finite dimensional $ \mathbb{C}$-vector space, and we can present $m_f$ by a matrix which is called the {\em action matrix} of $f$.  For any $i$, if we set $f=x_i$, then the eigenvalues of $m_{x_i}$ are the $x_i$-coordinates of the solutions of the system. Using these eigenvalues for each $i$, and  a test to verify whether or not a selection $n$-tuple of these eigenvalues vanishes the $f_i$'s, we could find the solutions of the system.  A more efficient way is to use eigenvectors. Let $f$ be a {\em generic} linear form in $A$, then we could read directly all solutions of the system from the right eigenvectors of $m_f$, see \cite{Cox}, page $64$. 
\subsection{Computation of final relative orientation}
After the resolution of the polynomial system, and the obtention of the parameters $T_x\ , T_y\ , T_z\ $ and $t$, it is possible to compute the finale relative orientation between the images.
If we suppose that $R_{ver1}$ is the rotation matrix defined in the section \ref{absoluOrientation}  for the image 1, and $R_{ver2}$ the same for the image 2, and $R_{\phi}$ the rotation matrix defined by $t$ (equation \ref{RotaY}), the final relative orientation between the images 1 and 2 is:
\begin{equation}
\begin{split}
&R_{final} = {R_{ver2}}^t\ R_{\phi}R_{ver1},\\
&\overrightarrow{T_{final}} = {R_{ver2}}^t\ \overrightarrow{T},\  where\ \overrightarrow{T} = [Tx, Ty, Tz]^t.
\end{split}
\label{OriFinal}
\end{equation}

\section{Experiments}\label{Experiments}
The accuracy of the relative orientation resolution, using a vertical vanishing point and 3 tie points, is based on three factors :\\
1- the accuracy of the polynomial resolution of the translation parameters $(Tx\ ,Ty\ ,Tz)$, and of the rotation around the $Y$ axis using the Gr\"obner bases,\\
2- the geometric accuracy for the estimation of the vertical direction,\\
3- the accuracy of the algorithm on tie points in presence of noise.\\
In order to evaluate the different impacts, we have in a first time worked on synthetic data in Section~\ref{sec:noise}, then we have used real data in Section \ref{sec:realperf}.

\subsection{Performance Under Noise}\label{sec:noise}
In this section, the performance of the 3 points method in noisy conditions has been studied and compared to the 5 points algorithm \cite{stewenius-engels-etal-ijprs-06} using the software provided by authors~\cite{Code5pointsST}.
The employed experimental setup is similar to \cite{Nister04}. The distance to the scene volume is used as the unit of measure, the baseline length being 0.3. The standard deviation of the noise is expressed in pixels of a 352x288 image as $\sigma = 1.0$. The field of view is equal to $45$ degrees. The depth varies between 0 to 2. Two different translation values have been treated, one in X (sideway motion) and one in Z (forward motion). The experiments involve 2500 random samples trials of point correspondences.
For each trial, we determinate the angle between estimated baseline and true baseline vector. This angle is called here translational error, and expressed in degrees. For the error estimation on the rotation matrix, the angle of $(R_{true}^TR_{estimate})$ is calculated, and the mean value for the  2500 random trials for each noise level is displayed.
From Figure \ref{fig:simuSimpleConfigTransX}, \ref{fig:simuSimpleConfigTransTransX}, \ref{fig:simuSimpleConfigTransZ} and \ref{fig:simuSimpleConfigTransTransZ}, we see that the 3-point algorithm is more robust to error caused by noise in sideway and forward motion for estimation of rotation and translation.

\begin{figure}[!ht]
   \centering
\includegraphics[width=0.7\linewidth]{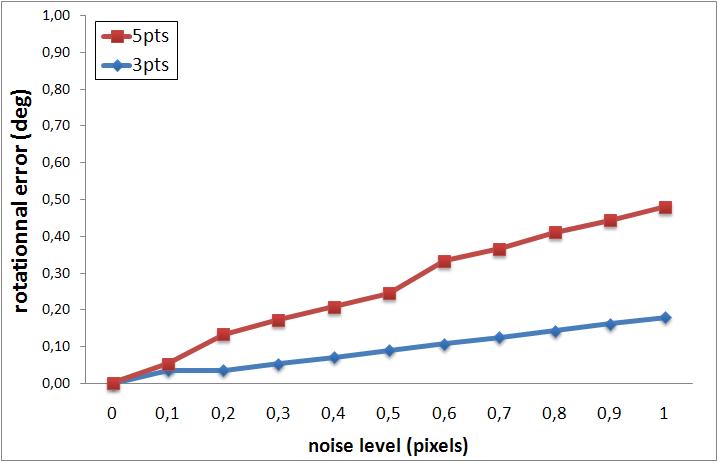}
    \caption{Error on the rotation (in degrees, sideway motion).}
\label{fig:simuSimpleConfigTransX}
\end{figure}

\begin{figure}[!ht]
   \centering
 \includegraphics[width=0.7\linewidth]{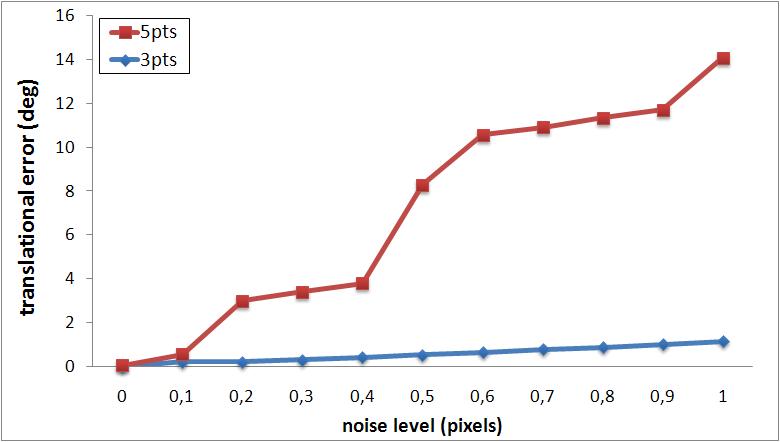}
    \caption {Error on the baseline orientation (in degrees, sideway motion).}
\label{fig:simuSimpleConfigTransTransX}
\end{figure}

\begin{figure}[!ht]
   \centering
\includegraphics[width=0.7\linewidth]{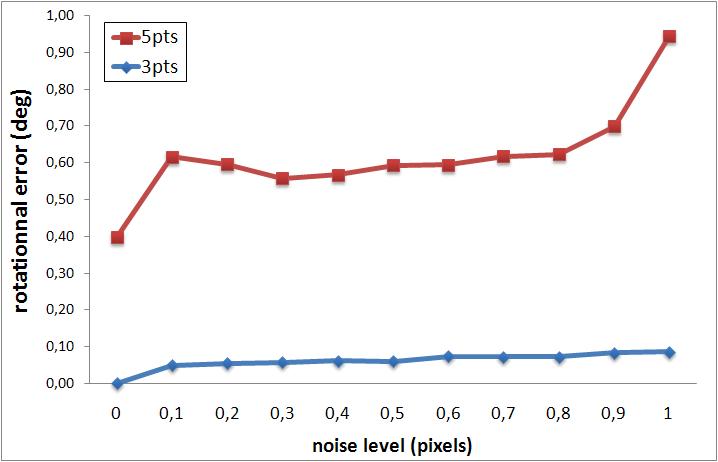}
    \caption{Error on the rotation (in degrees, forward motion)}
\label{fig:simuSimpleConfigTransZ}
\end{figure}

\begin{figure}[!ht]
   \centering
 \includegraphics[width=0.7\linewidth]{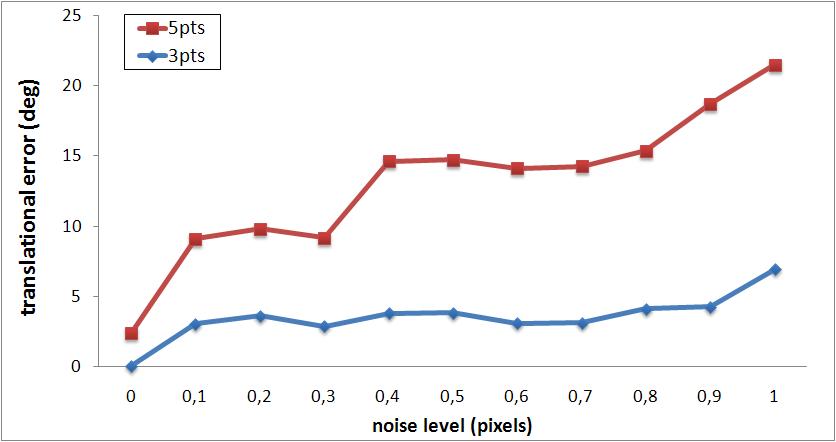}
    \caption{Error on the baseline orientation (in degrees, forward motion).}
\label{fig:simuSimpleConfigTransTransZ}
\end{figure}

Now let us compare 3-point and five-point algorithm on a planar scene. In this configuration all the points of the scene in the world have the same $Z$ (here equal to 2).
The results for the estimation of the rotation (Figure~\ref{fig:simuSimpleConfigTransXPlane}) show that the two algorithms provide a good determination of the rotation, but the 3-point gives much better results than the 5-point one for the base determination in sideway motion (Figure \ref{fig:simuSimpleConfigTransTransXPlane}). This weakness of the 5-point algorithm in planar scene has been discussed in \cite{Segvic}.
\begin{figure}[!ht]
   \centering
\includegraphics[width=0.7\linewidth]{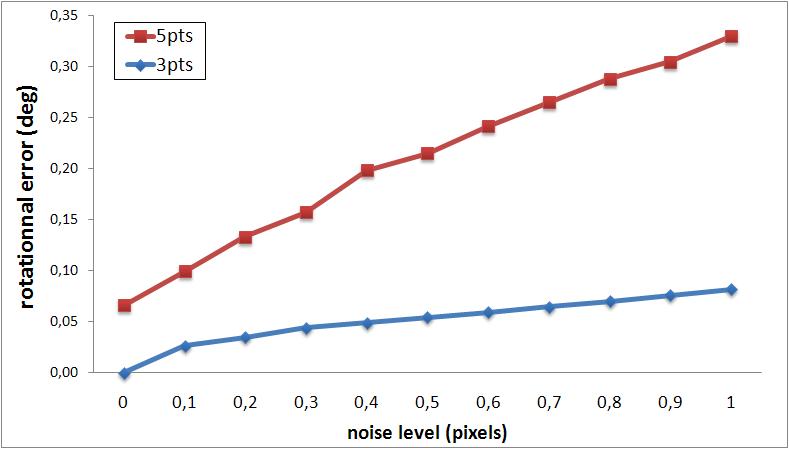}
  \caption{Error on the rotation (in degrees) in planar configuration (sideway motion) }.
\label{fig:simuSimpleConfigTransXPlane}
\end{figure}

\begin{figure}[!ht]
   \centering
  \includegraphics[width=0.7\linewidth]{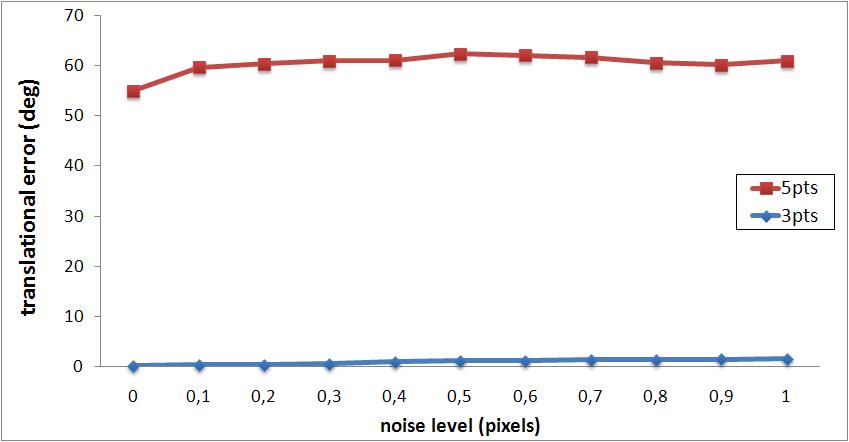}
  \caption{Error on the base orientation (in degrees) in planar configuration (sideway motion)}
\label{fig:simuSimpleConfigTransTransXPlane}
\end{figure}

\begin{figure}[!ht]
   \centering
\includegraphics[width=0.7\linewidth]{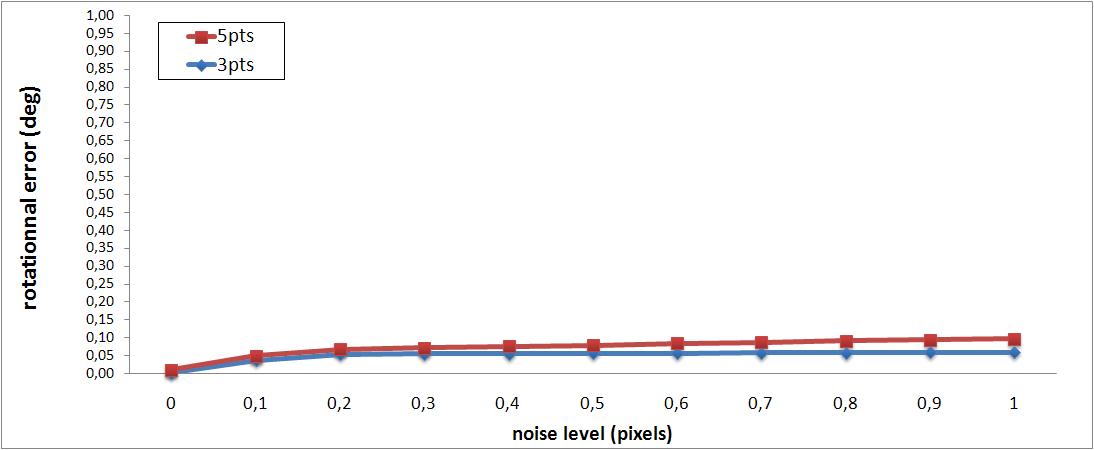}
  \caption{ Error on the rotation (in degrees) in planar configuration (forward motion).}
\label{fig:simuSimpleConfigTransZPlane}
\end{figure}

\begin{figure}[!ht]
   \centering
    \includegraphics[width=0.7\linewidth]{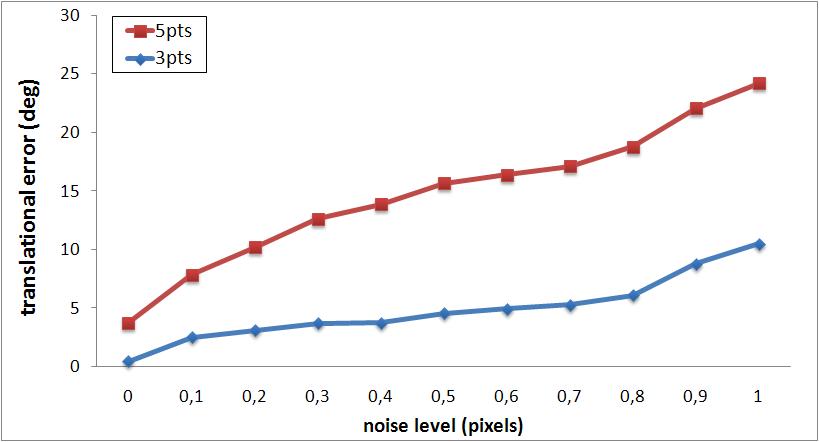}
  \caption{Error on the base orientation (in degrees) in planar configuration (forward motion)}
\label{fig:simuSimpleConfigTransTransZPlane}
\end{figure}

\subsubsection{Impact of the accuracy of the vertical direction on the estimation of relative orientation}
We have introduced an error of $0$ to $0.5\,^{\circ}$ on the angular accuracy of the vertical direction.
Today for example, a low-cost inertial sensor such as Xsens-MTi~\cite{MTI} gives a precison around $0.5\,^{\circ}$ on rotation angle around X axis and Z axis (the vertical direction being Y axis). Of course, some high accuracy IMU are available, they may reach an accuracy better than $0.01\,^{\circ}$ on the orientation angles if properly coupled with other sensors (e.g. GPS).
Using an automatic vanishing point detection specially in urban scene, we get a very precise vertical direction (better than $0.001\,^{\circ}$), as it will be shown later.
We have checked the impact of this accuracy on the determination of the rotation and the base. 
(Figure~\ref{fig:simuErrorVPTX} and Figure~\ref{fig:simuErrorVPTZ}).

\begin{figure}[!ht]
   \centering
   \begin{tabular}{cc}
\includegraphics[width=0.45\linewidth]{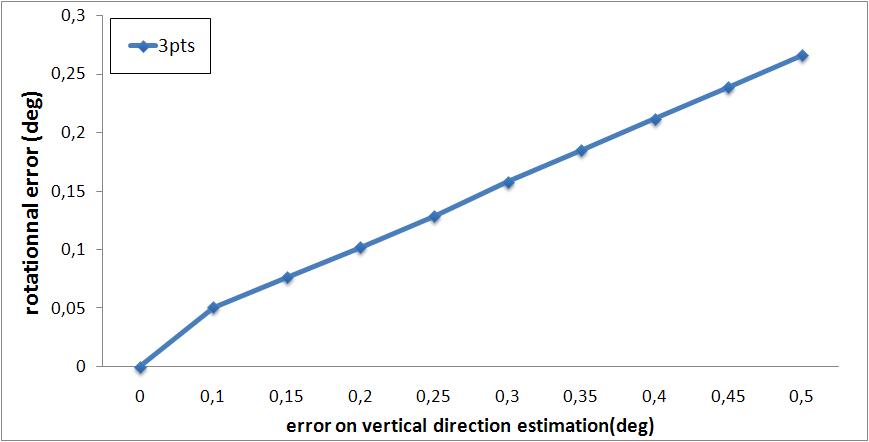}&      \includegraphics[width=0.45\linewidth]{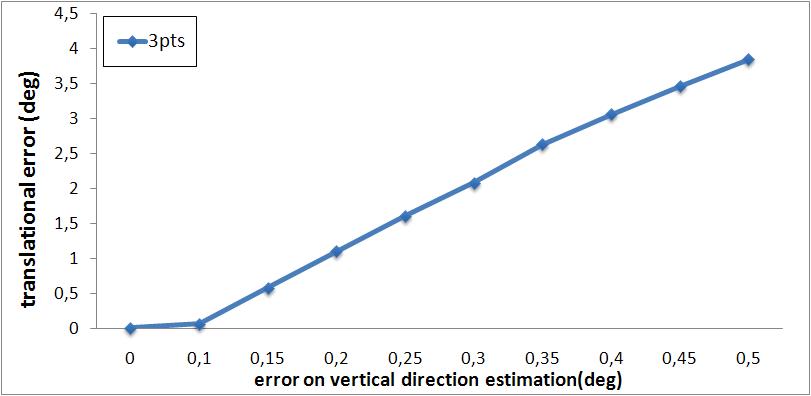}\\
(a) & (b)\\
 \end{tabular}
   \caption{Impact of the geometric accuracy of the vertical direction on the estimation of a) the rotation (in degrees), and b) the base orientation (in degrees) in sideway motion.}
\label{fig:simuErrorVPTX}
\end{figure}

\begin{figure}[!ht]
   \centering
   \begin{tabular}{cc}
\includegraphics[width=0.45\linewidth]{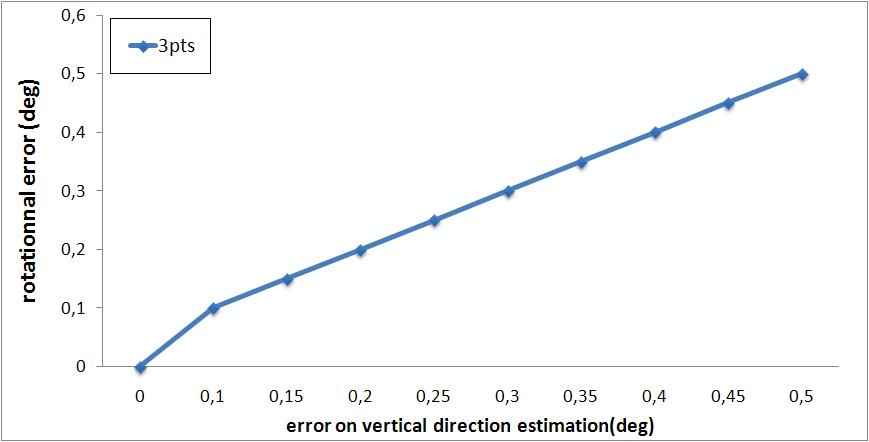}&      \includegraphics[width=0.45\linewidth]{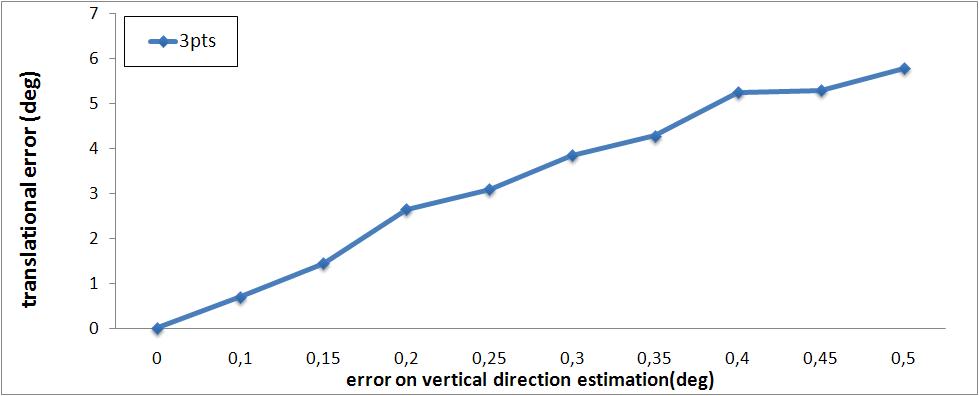}\\
(a) & (b)\\
 \end{tabular}
  \caption{Impact of the geometric accuracy of the vertical direction on the estimation of a) the rotation (in degrees), and b) the base orientation (in degrees) in forward motion.}
\label{fig:simuErrorVPTZ}
\end{figure}

\subsection{Real Example}\label{sec:realperf}
So as to provide a numerical example on real images, we have chosen to work on the 9-images sequence "entry-P10" of the online database ~\cite{EPFLdatabase}. In this database we know all the intrinsec and external parameters.
First, we extracted the vanishing points on each image.
We used the algorithm of \cite{Kalantari} because beyond its high speed, it allows an error propagation on the vanishing points according to the error on the segments detection.
We express this error in an angular manner. The results of the angular errors are shown in the table 1. As one can see it, the determination of the vertical vanishing point is very precise and according to the Figure \ref{fig:simuErrorVPTX} and \ref{fig:simuErrorVPTZ} it induced an error close to zero.
\begin{table}
\label{tab:ErreurPDF}
\begin{center}
\begin{tabular}{|l|c|}
\hline
Image & Angular error on vertical direction in degree \\
\hline\hline
0000 & 0.002569 \\
0001 & 0.0066\\
0002 & 0.001584\\
0003 & 0.001443\\
0004 & 0.000899 \\
0005 & 0.00115\\
0006 & 0.001445 \\
0007 & 0.005018 \\
0008 & 0.002424\\
0009 & 0.002223\\
\hline
\end{tabular}
\end{center}
\caption{Results. Vertical direction detection using the vertical vanishing point.}
\end{table}
Then, we have computed the relative orientation for 3 successive images (each time, 2 following couples of images). The interest points are extracted using SIFT ~\cite{Sift} algorithm.
The results are presented in the Figure \ref{fig:orientationresultEPFL}. The mean value of angular errors on the rotation amounts to $0.82$ degree. For the estimation of the translation, this error amounts to $1.33$ degree. These results show clearly the efficiency and robustness of the method.

\begin{figure}[!ht]
\begin{center}
\includegraphics[width=1\linewidth]{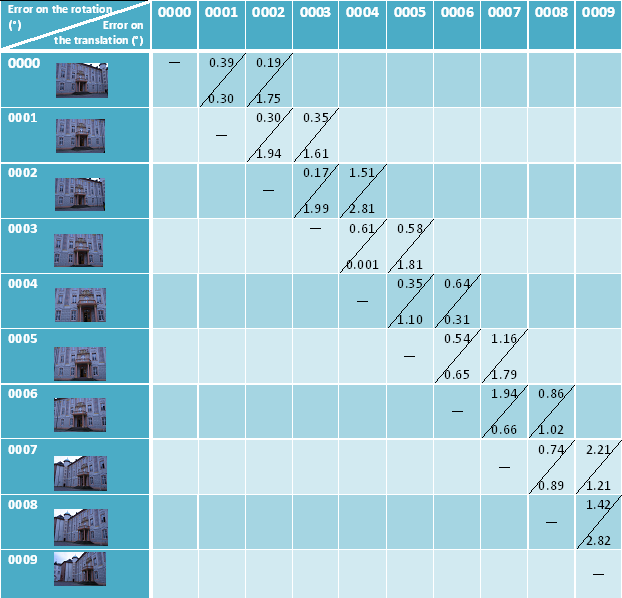}
\end{center}
  \caption{Result on "entry-P10" sequence. Each cell contiens the error on rotation in degrees (upper left) and error on the translation in degrees (bottom right).}
\label{fig:orientationresultEPFL}
\end{figure}

\subsection{Time Perfomance}
The resolution of the polynomial system and detection of vanishing point was written in C ++. With a 1.60 GHz PC the time of each resolution is about $2\ \mu s$, allowing real-time application. We may note that the selection process using RanSac \cite{RanSac} among the SIFT points is running considerably faster on 3-point than on 5-point algorithm.
\section{Summary and Conclusions}
Today, more and more low-cost personal devices include MEMS-IMU in complement to cameras, these devices allow to provide very easily the direction of the vertical in the image. Furthermore, image based automatic extraction of the vertical vanishing point offers a very high accuracy alternative, if needed. So, here, we have demonstrated the advantage of using the vertical direction, and an efficient algorithm for solving the relative orientation problem with this information has been presented. In addition to a considerable acceleration, compared with the classical 5 point solution, our algorithm provide a noticeable accuracy improvement for the baseline  estimation. Another interesting feature improvement has been demonstrated: the planar scenes raise no more problem in baseline estimation. This advantageous result is due to an appropriate problem formulation using in a explicit way the significant parameters of the relative orientation (parameters of the rotation and the translation).
{\small

}

\end{document}